\documentclass[a4paper,12pt]{article}
\usepackage[top=2cm,bottom=2cm,left=2cm,right=2cm]{geometry}

\usepackage{graphicx}
\usepackage{subcaption}
\usepackage{hyperref}
\usepackage{titlesec}
\usepackage{enumitem}
\usepackage{booktabs}
\usepackage{varwidth}
\usepackage{tasks}
\usepackage{booktabs}
\usepackage{xcolor}
\usepackage{pdflscape}
\usepackage{longtable}
\usepackage{array}
\usepackage{amsmath}

\usepackage{float}
\usepackage{tabularx}

\usepackage{siunitx}

\newcommand{\ok}[1]{{\color{green!55!black}\underline{{\color{black}#1}}}}
\newcommand{\bad}[1]{{\color{red!75!black}\underline{{\color{black}#1}}}}
\newcommand{\miss}[1]{{\color{orange!85!black}\underline{{\color{black}#1}}}}

\usepackage[dvipsnames]{xcolor}
\usepackage[most]{tcolorbox}
\tcbuselibrary{fitting}

\usepackage[super,sort&compress,comma]{natbib}
\usepackage[font=footnotesize,labelfont={bf,it}, textfont=it]{caption}
\usepackage[labelsep=period]{caption}

\usepackage{pdflscape}
\usepackage{fancyhdr} 
\usepackage{dirtytalk}

\fancypagestyle{mylandscape}{
\fancyhf{} 
\fancyfoot{
\makebox[\textwidth][r]{
\rlap{\hspace{.75cm}
\smash{
  \raisebox{4.87in}{
    \rotatebox{90}{\thepage}}}}}}
}

\newcommand{\keywords}[1]{%
  \vspace{0.5em}
  \noindent\textbf{Keywords:} #1
}

\begin{document}

\title{Automatic Ordinary Differential Equations Discovery For Biological Systems Using Large Language Model Powered Agentic System}

\author{David Krongauz\(^{1,2}\), Arad Zulti \(^{1,2}\), Eran Segal\(^{1,3}\), Teddy Lazebnik\(^{4,5,*}\) \\\\ \(^1\) Department of Computer Science and Applied Mathematics, \\Weizmann Institute of Science, Rehovot, Israel \\ \(^2\) Department of Molecular Cell Biology,\\ Weizmann Institute of Science, Rehovot, Israel \\ \(^3\) Mohamed bin Zayed University of Artificial Intelligence,\\ Abu Dhabi, United Arab Emirates \\ \(^4\) Department of Information Systems, University of Haifa, Haifa, Israel \\ \(^5\) Department of Computing, Jonkoping University, Jonkoping, Sweden \\ \(^*\) Corresponding author: \url{teddy.lazebnik@ju.se}}

\date{ }

\maketitle

\begin{abstract} 
\noindent
Automatic scientific discovery has long been a goal of computational scholars - a machine that can discover nature's secrets on its own, moving computational systems beyond data-fitting tools toward the generation and refinement of mechanistic models of the universe. Recent advances in symbolic regression (SR) and large-language-model (LLM)-based agents suggest that such systems can recover equations from data, incorporate domain priors, and automate parts of the research workflow. However, most existing approaches either focus on narrow equation-discovery benchmarks or broad end-to-end automation pipelines, while biological systems remain comparatively underexplored. Here, we introduce the \textsc{MEDA} system, an LLM- and SR-powered agentic framework for discovering ordinary-differential-equation (ODE) models of biological and biologically inspired dynamical systems. \textsc{MEDA} retrieves background knowledge, defines admissible variables, generates mechanistic constraints, proposes candidate ODEs, and fits and evaluates them. We evaluate it across canonical model retrieval, reasoning-based extrapolation to unseen variants, and open-ended discovery, with and without experimental data. Across these settings, \textsc{MEDA} recovered the correct state variables, achieved strong structural recovery in retrieval and extrapolation tasks, and produced biologically plausible discovery-oriented models. Ablation and robustness analyses show that knowledge-guided formalization and mechanistic constraints are load-bearing components, whereas numerical fitting alone can preserve trajectory-compatible but biologically incorrect equations.  
\end{abstract}
\newpage

\section{Main}
Automatic scientific discovery (ASD) has sparked the imagination of scientists across space and time, pushing a wide and long chain of studies to achieve this ultimate goal \cite{lu2026towards,sparkes2010towards,lenat11984automated,langley1987scientific}. ASD has moved from a longstanding aspiration of artificial intelligence (AI) into an active research program spanning equation discovery, closed-loop experimentation, and increasingly agentic workflows \cite{kramer2023automated,xin2025agentic,merchant2023scaling}. In parallel, large language models (LLMs) are being positioned not merely as interfaces for scientific text, but as components of the scientific method itself, with proposed roles in background synthesis, hypothesis generation, experimental planning, and interpretation \cite{zhang2025role,faldor2025omni,girotra2023ideas,hu2025automated}. This momentum has created a renewed expectation that parts of scientific reasoning may become automatable, or at least systematically augmentable, across a wide range of domains \cite{vinuesa2026decoding,messeri2026uncritical,that2026ai}.

Indeed, high-visibility systems now attempt end-to-end research automation, including ideation, code generation, experimentation, visualization, and paper writing \cite{lu2024aiscientist,schmidgall2025agentlab}. Nevertheless, the current wave of LLM-driven scientific automation often over-promises relative to its empirical reliability \cite{narayanaswamy2023can,song2026paperorchestra}. For example, on memorization-resistant equation-discovery benchmarks, the strongest reported systems still recover the correct symbolic form in only a minority of cases \cite{shojaee2025llmsrbench,bideh2026mdbench,brum2026discovering}. Likewise, when used as scientific verifiers, state-of-the-art LLMs exhibit poor recall and precision on real errors in published manuscripts \cite{son2025spot}. These findings suggest that fluent generation of reports or academic papers should not be conflated with successful scientific discovery. If the goal is to automate central scientific work, then the target should not be the production of polished manuscripts, but the construction, criticism, and refinement of mechanistic hypotheses \cite{kumaran2026competing,fu2026multiple}.

This distinction is especially important for exact and life sciences that are based on mathematical formalization to describe the world \cite{cahan2023conversation,ahmed2023fda,chen2024applying,guo2024diffusion}, often using ordinary differential equations (ODEs) based modeling \cite{giampiccolo2024robust,luders2022odebase,vieira2022computational}. Across systems biology, epidemiology, ecology, and population dynamics, ODEs remain one of the most compact and useful languages for expressing mechanistic assumptions about rates, interactions, feedback, and intervention effects. The main challenge is structural rather than merely numerical, as one must infer not only parameter values, but also which terms should appear in the equations at all. Symbolic regression (SR) addresses this problem by searching jointly over equation structure and coefficients, rather than fitting parameters inside a prespecified functional family \cite{makke2024srreview,schmidt2009distilling}. For instance, given a time series of predator and prey abundances, SR may discover that a bilinear interaction term such as $xy$ is required in the governing law, rather than merely approximating the trajectories with a black-box predictor \cite{brum2026discovering}. In spite of that, it is inherently solving only the \say{last mile} challenge of discovering symbolic models from numerical data, neglecting most of the scientific process of feature definition, dynamics inference from partial observations, and even evaluation of the obtained model's usability as a scientific tool \cite{wang2023scientific,chigbu2023science}. 

To this end, recent attempts have partially aimed to address these challenges using an optimization loop of human (or scientist) feedback \cite{scheurer2025role,kallmes2025human,lee2020human}. For instance, in ecology, SR has already been framed as a collaborative human--machine process for surfacing relationships, models, and candidate principles from data \cite{cardoso2020ecology}. In equation discovery more broadly, SciMED showed that explicit scientist-in-the-loop design can inject domain knowledge into symbolic search so as to improve robustness, plausibility, and interpretability \cite{keren2023scimed}. Other lines of work have demonstrated that constraining the hypothesis space through sparse libraries or grammar-based priors can substantially improve the recovery of governing equations \cite{brunton2016sindy,brence2021grammars}. More recently, LLM-SR illustrated that LLMs can contribute scientific priors and programmatic scaffolds to symbolic search \cite{shojaee2025llmsr} or even guide the SR search process as a judge \cite{taskin2026knowledge}, but it still primarily treats equation discovery as a benchmarked search problem rather than as a broader scientific reasoning problem cycle.  

In this study, we deliberately reduce the objective of ASD from automatically writing academic papers, with all the sub-tasks related to such a process, to the more central scientific task of discovering ODEs together with explicit mechanistic reasons for them, which further extends recent SR systems' scope. While most of the knowledge-integrated SR models focus on physics \cite{shmuel2026interpretable,zhang2025physics,udrescu2020ai,kubalik2021multi}, where one can express rules using formal equations \cite{lu2016using} and take advantage of prior knowledge in the form of symmetries \cite{kronberger2022shape}, or general-purpose reduction forms (such as Newton's laws) \cite{fox2024incorporating,kammerer2020symbolic}, in this study, we focused on biological use-cases, which is a more challenging domain lacking such properties but do have a rich history of ODE modeling \cite{rihan2021delay,ilea2012ordinary,gennemark2007efficient}. Specifically, biological ODE discovery is particularly difficult because the data are noisy, sparse, heterogeneous, and often only partially informative, while prior knowledge is typically incomplete rather than absent \cite{wu2025noisybiology,egan2024odesparse}. Our LLM-powered agentic-based, therefore, begins by searching for relevant background knowledge, then schematically defines admissible dynamical structures through biologically motivated constraints, and only then generates, fits, and critiques candidate equations. Candidate hypotheses are evaluated not only by numerical agreement with observed dynamics, if such is provided, but also by conceptual criteria such as mechanistic consistency, admissible interactions, and interpretability, all inferred implicitly by the model rather than by an expert user. By coupling SR with knowledge-enriched numerical fitting and iterative scientific critique, the proposed framework aims both to reconstruct known biological ODEs and to discover new ODEs when the literature provides only partial mechanistic guidance.

\section{Mechanistic Equation Discovery Agentic System}
\label{sec:mendel}
\textsc{MEDA} (\emph{Mechanistic Equation Discovery Agentic}) system discovers ODE models of mechanistic systems by coupling LLM agents with constrained SR. Rather than fitting trajectories with a black-box predictor, or asking an LLM to output an equation directly, \textsc{MEDA} reconstructs the scientific reasoning around a model. It follows a path similar to the process of a human scientist as it gathers background knowledge from the literature, defines the admissible state variables and biological constraints, searches for sparse ODE systems that satisfy them, and then critiques and evaluates the result. Each step is performed by a specialized agent that reads and writes a human-readable artifact, so the full chain of reasoning behind a discovered equation can be inspected, interpreted, and validated by a human expert. 

\begin{figure}[!ht]
\centering
\includegraphics[width=\textwidth]{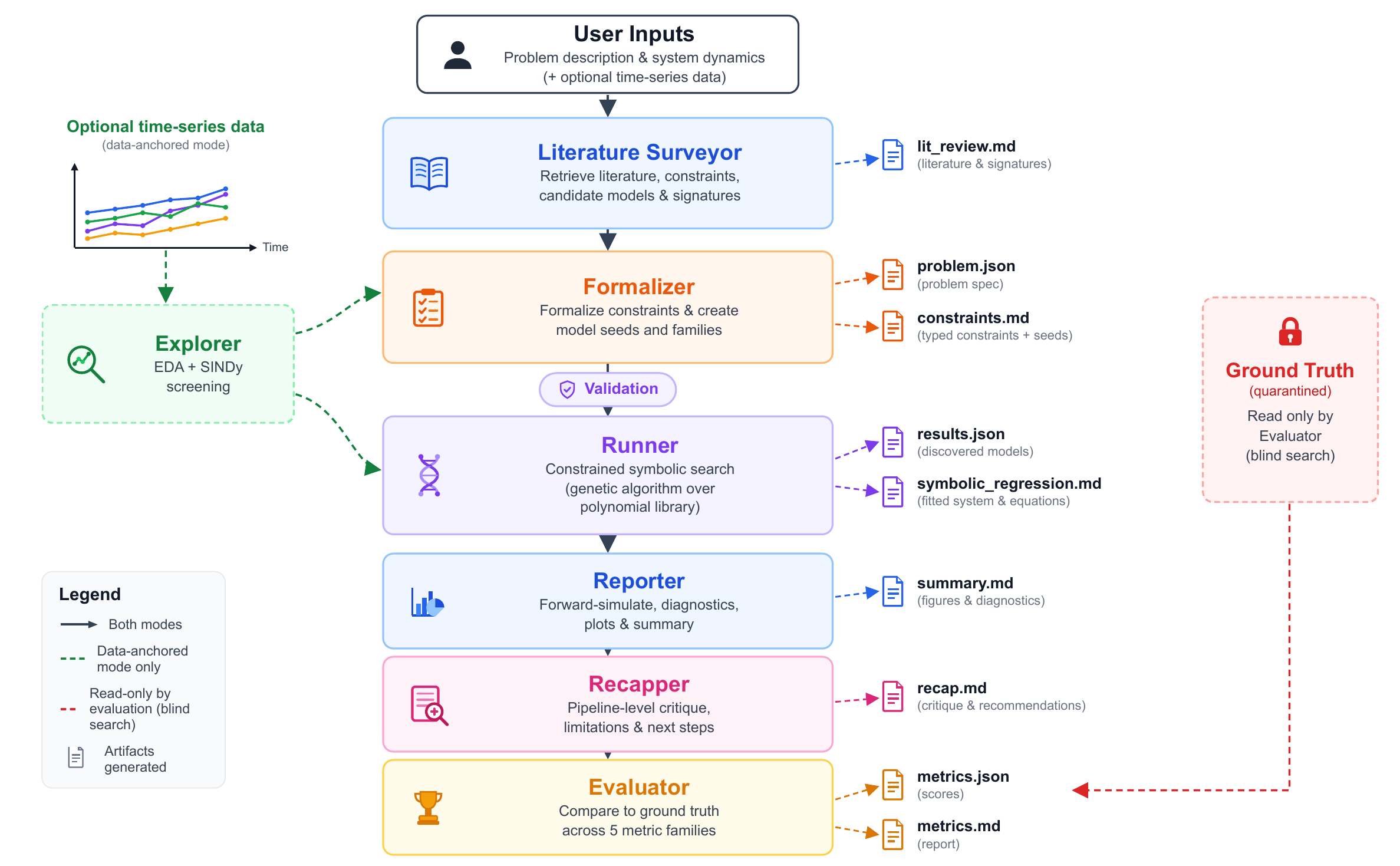}
\caption{Architecture of \textit{MEDA}. Specialized agents (centre) each read a defined input artifact and write a defined output artifact (right) to a shared session directory. Solid arrows apply in both operating modes; green dashed arrows are active only in data-anchored mode, in which a supplied time series feeds the \textbf{Explorer} (EDA + SINDy screening) and the \textbf{Runner}. The ground-truth equations are quarantined from all discovery agents and read only after the blind search ends, by the \textbf{Evaluator} (red dashed arrow).}
\label{fig:architecture}
\end{figure}

Figure~\ref{fig:architecture} provides a schematic view of \textit{MEDA}'s architecture. The pipeline proceeds in stages. A \textit{Literature} Surveyor compiles cited background on the target system, including expected constraints, candidate model families, and qualitative dynamical signatures. A \textit{Formalizer} curates this material into a machine-readable problem specification: it selects the relevant constraints, expresses them in a typed grammar, and proposes seed equations. A \textit{Runner} then performs an evolutionary search over polynomial ODE systems, scoring candidates by how well they satisfy the biological constraints, how sparse they are, and, when data are available, how well they reproduce the observed trajectories. Finally, a \textit{Reporter}, \textit{Recapper}, and \textit{Evaluator} summarize, critique, and (against held-out ground truth) score the discovered model.  

\textit{MEDA} is organized as a set of specialized agents coordinated by an orchestrator. Each agent reads a defined input artifact, performs a narrow scientific or computational role, and writes a structured output artifact to a shared session directory. This design makes the discovery process inspectable: the literature review, data summary, formal constraints, seed equations, selected ODE system, diagnostics, critique, and evaluation metrics can each be audited independently. The system operates in two modes. In constraint-only mode, no empirical time series is supplied, and the search is guided by literature-derived biological constraints, qualitative dynamical signatures, and a sparsity preference. In data-anchored mode, the user additionally provides a time series; the system summarizes the trajectories, optionally performs a SINDy-like sparse-regression screen, and augments the search objective with a trajectory-fit score. In both modes, ground-truth equations used for benchmarking are quarantined from all discovery agents and are accessed only after model generation by the post-hoc recapping and evaluation stages. The pipeline proceeds through a sequence of interpretable artifacts. A Data Explorer, used only in data-anchored mode, characterizes the supplied trajectories and identifies candidate data-supported terms. A Literature Surveyer summarizes relevant biological mechanisms, constraints, canonical model families, and qualitative signatures. A Formalizer converts this information into a machine-readable problem specification containing state variables, typed hard and soft constraints, candidate seed equations, protected core terms, polynomial-degree limits, and constraint weights. A validator checks this specification before search begins. The Runner then performs constrained symbolic search over sparse polynomial ODE systems using a composite objective that combines constraint satisfaction, data fit when available, and sparsity, while hard-constraint gating and core-term protection preserve biological admissibility. Finally, the Reporter translates the selected model into human-readable scientific output, the Recapper diagnoses system-level failure modes, and the Evaluator compares the result with ground truth when available. Full technical details and default implementation settings are provided in the Appendix.

Importantly, \textit{MEDA} runs in two modes. In \say{constraint-only} mode no measurements are provided, and the search is driven purely by literature-derived constraints and a preference for parsimony; in \say{data-anchored} mode the user additionally supplies a time series, which a \textit{Data Explorer} characterizes and the Runner uses as an extra fit objective. In both modes, any ground-truth equations used for benchmarking are quarantined from the discovery agents and read only afterwards, so the search remains blind. Due to the fact that candidates are judged by mechanistic plausibility and not numerical fit alone, \textit{MEDA} can propose defensible models even when the literature offers only partial guidance and no data are available. We provide \textit{MEDA} as an online tool. 

\section{Experimental results}
We evaluated model construction using a hierarchy of metrics that follows the structure of the \textit{MEDA} pipeline. Variable precision, recall and \(F_1\) measure whether the system identified the correct dynamical state space. This metric reveals if \textit{MEDA} pick the right variables of the dynamic, operating similarly to a feature selection/engineering phase. Next, constraint precision, recall and \(F_1\) assess whether the intermediate biological restrictions were consistent with the reference mechanism. Since the constraints are the core of the equation design, these metrics allow to evaluate the equation reconstruction under the model's discovery. Finally, term precision, recall and \(F_1\) quantify recovery of the symbolic building blocks of the ODEs, independent of exact coefficient values. To capture whole-expression similarity, we report an expression score defined as \(1-\mathrm{TED}\), where TED is the normalized Tree Edit Distance between the predicted and reference expressions \cite{taskin2026knowledge}. We also report an inclusion score, which measures whether the reference terms are contained in the generated expression, and an expert plausibility score, which evaluates whether the complete model is biologically coherent even when exact symbolic recovery is not unique. The plausibility score is computed to be the average of five scores between zero and one given by five biomathematicians, each with a PhD in mathematics and over ten years of academic experience as prime investigators.

Table \ref{tab:result1} provides a term-level audit of the generated ODEs under two inference regimes: construction from biological constraints alone and construction with additional fitting data. In the constraints-only rows, numerical coefficients are represented by symbolic coefficients (\(c_1,c_2,\ldots\)), since the comparison focuses on whether the correct mechanistic terms were selected rather than on parameter calibration, as data fitting is lacking in this configuration. Green-underlined terms denote recovered reference terms with the correct sign, red-underlined terms denote spurious or sign-inconsistent terms, and orange-underlined terms denote reference terms that were absent from the prediction. Several patterns emerge from the table. First, the system reliably identifies the relevant state variables and often recovers the correct mechanistic skeleton even when the final symbolic expression is imperfect, indicating that most failures occur at the level of term selection rather than problem formulation. Second, fitting data are beneficial when the remaining ambiguity is primarily a search or calibration problem: for example, Lotka--Volterra and misinformation dynamics improve to exact or near-exact structural recovery with data. However, the effect of data is not monotonic. In FitzHugh--Nagumo adaptation and chronic wound healing, data-driven fitting favours trajectory-compatible but mechanistically incorrect expressions, introducing high-degree or spurious terms while missing defining biological couplings. The host--dual-pathogen case exposes a different challenge: the same wrong biological family persists with and without data, suggesting that errors introduced by constraint generation cannot always be corrected downstream by numerical fitting.

\begin{landscape}
\tiny
\setlength{\tabcolsep}{2.5pt}
\renewcommand{\arraystretch}{1.20}

\begin{longtable}{%
p{0.060\linewidth}
p{0.030\linewidth}
p{0.130\linewidth}
p{0.365\linewidth}
cccccccc}
\caption{Per-case performance with and without fitting data.}
\label{tab:result1}\\

\toprule
\textbf{Data} & \textbf{Setting} & \textbf{Problem} & \textbf{Predicted equation}
& \textbf{Var. F$_1$}
& \textbf{Constr. F$_1$}
& \textbf{Term P}
& \textbf{Term R}
& \textbf{Term F$_1$}
& \textbf{Expr. score}
& \textbf{Inclusion}
& \textbf{Plaus.} \\
\midrule
\endfirsthead

\toprule
Data & Setting & Problem & Predicted equation
& Var. F$_1$
& Constr. F$_1$
& Term P
& Term R
& Term F$_1$
& Expr. score
& Inclusion
& Plaus. \\
\midrule
\endhead

\midrule
\multicolumn{12}{r}{Continued on next page}
\endfoot

\bottomrule
\endlastfoot

No data 
& IR 
& Single population growth
& $\displaystyle \frac{dN}{dt} =
\ok{+c_1N}
\ok{-c_2N^2}$
& 1.00 & 1.00 & 1.00 & 1.00 & 1.00 & 1.00 & 1.00 & 1.00 \\

With data 
& IR 
& Single population growth
& $\displaystyle \frac{dN}{dt} =
\ok{+0.23N}
\ok{-0.0001N^2}$
& 1.00 & 0.91 & 1.00 & 1.00 & 1.00 & 1.00 & 1.00 & 1.00 \\

\midrule

No data 
& IR 
& Lotka--Volterra predator--prey
& $\displaystyle \frac{dH}{dt} =
\ok{+c_1H}
\ok{-c_2HL}
\bad{-c_3H^2}$

$\displaystyle \frac{dL}{dt} =
\ok{+c_4HL}
\ok{-c_5L}
\bad{-c_6L^2}$
& 1.00 & 1.00 & 0.67 & 1.00 & 0.80 & 0.75 & 1.00 & 0.93 \\

With data 
& IR 
& Lotka--Volterra predator--prey
& $\displaystyle \frac{dx}{dt} =
\ok{+0.80x}
\ok{-0.10xy}$

$\displaystyle \frac{dy}{dt} =
\ok{+0.10xy}
\ok{-0.50y}$
& 1.00 & 1.00 & 1.00 & 1.00 & 1.00 & 1.00 & 1.00 & 1.00 \\

\midrule

No data 
& IR 
& SIR epidemic
& $\displaystyle \frac{dS}{dt} =
\ok{-c_1SI}$

$\displaystyle \frac{dI}{dt} =
\ok{+c_2SI}
\ok{-c_3I}$

$\displaystyle \frac{dR}{dt} =
\ok{+c_4I}$
& 1.00 & 1.00 & 1.00 & 1.00 & 1.00 & 1.00 & 1.00 & 1.00 \\

With data 
& IR 
& SIR epidemic
& $\displaystyle \frac{dS}{dt} =
\ok{-0.0003SI}$

$\displaystyle \frac{dI}{dt} =
\ok{+0.0003SI}
\ok{-0.10I}$

$\displaystyle \frac{dR}{dt} =
\ok{+0.10I}$
& 1.00 & 1.00 & 1.00 & 1.00 & 1.00 & 1.00 & 1.00 & 0.98 \\

\midrule

No data 
& IR 
& Allee fixed threshold
& $\displaystyle \frac{dN}{dt} =
\ok{-c_1N}
\ok{+c_2N^2}
\ok{-c_3N^3}$
& 1.00 & 0.86 & 1.00 & 1.00 & 1.00 & 1.00 & 1.00 & 0.92 \\

With data 
& IR 
& Allee fixed threshold
& $\displaystyle \frac{dN}{dt} =
\ok{-0.46N}
\ok{+0.03N^2}
\ok{-0.0002N^3}$
& 1.00 & 0.92 & 1.00 & 1.00 & 1.00 & 1.00 & 1.00 & 0.99 \\

\midrule

No data 
& IR 
& FitzHugh--Nagumo
& $\displaystyle \frac{dV}{dt} =
\ok{+c_1V}
\ok{-c_2V^3}
\ok{-c_3W}
\quad [\text{\scriptsize missing: } \miss{+c_4}]$

$\displaystyle \frac{dW}{dt} =
\ok{+c_5V}
\ok{-c_6W}
\quad [\text{\scriptsize missing: } \miss{+c_7}]$
& 1.00 & 0.89 & 1.00 & 0.71 & 0.83 & 0.75 & 0.75 & 0.94 \\

With data 
& IR 
& FitzHugh--Nagumo
& $\displaystyle \frac{dv}{dt} =
\ok{+0.33v}
\ok{-0.16v^3}
\ok{-0.44w}
\bad{+0.17v^2}
\quad [\text{\scriptsize missing: } \miss{+I}]$

$\displaystyle \frac{dw}{dt} =
\ok{+0.10v}
\ok{-0.003w}
\quad [\text{\scriptsize missing: } \miss{+c}]$
& 1.00 & 0.86 & 0.83 & 0.71 & 0.77 & 0.75 & 0.75 & 0.78 \\

\midrule

No data 
& EX 
& SIR age-structured
& $\displaystyle \frac{dS_y}{dt} =
\ok{-c_1S_yI_y}
\ok{-c_2S_yI_o}$

$\displaystyle \frac{dI_y}{dt} =
\ok{+c_3S_yI_y}
\ok{+c_4S_yI_o}
\ok{-c_5I_y}$

$\displaystyle \frac{dR_y}{dt} =
\ok{+c_6I_y}$

$\displaystyle \frac{dS_o}{dt} =
\ok{-c_7S_oI_y}
\ok{-c_8S_oI_o}$

$\displaystyle \frac{dI_o}{dt} =
\ok{+c_9S_oI_y}
\ok{+c_{10}S_oI_o}
\ok{-c_{11}I_o}$

$\displaystyle \frac{dR_o}{dt} =
\ok{+c_{12}I_o}$
& 1.00 & 1.00 & 1.00 & 1.00 & 1.00 & 1.00 & 1.00 & 0.96 \\

With data 
& EX 
& SIR age-structured
& $\displaystyle \frac{dI_Y}{dt} =
\ok{+I_YS_Y}
\ok{+I_OS_Y}
\ok{-I_Y}$

\textit{Same equation}
& 1.00 & 1.00 & 1.00 & 1.00 & 1.00 & 1.00 & 1.00 & 0.94 \\

\midrule

No data 
& EX 
& Host dual pathogen
& $\displaystyle \frac{dN}{dt} =
\ok{+c_1N}
\ok{-c_2N^2}
\ok{-c_3NP_1}
\ok{-c_4NP_2}$

$\displaystyle \frac{dP_1}{dt} =
\bad{-c_5P_1}
\ok{-c_6P_1P_2}
\bad{+c_7NP_1}$

$\displaystyle \frac{dP_2}{dt} =
\bad{-c_8P_2}
\ok{-c_9P_1P_2}
\bad{+c_{10}NP_2}$
& 1.00 & 0.93 & 0.80 & 1.00 & 0.89 & 0.78 & 1.00 & 0.87 \\

With data 
& EX 
& Host dual pathogen
& $\displaystyle \frac{dN}{dt} =
\ok{+0.25N}
\ok{-0.0001N^2}
\ok{-0.019NP}
\ok{-0.001NQ}$

$\displaystyle \frac{dP}{dt} =
\bad{-0.015P}
\ok{-0.010PQ}
\bad{+0.000004NP}$

$\displaystyle \frac{dQ}{dt} =
\bad{-0.029Q}
\ok{-0.023PQ}
\bad{+0.000001NQ}$
& 1.00 & 0.77 & 0.80 & 1.00 & 0.89 & 0.78 & 1.00 & 0.80 \\

\midrule

No data 
& EX 
& IGP 4-species
& $\displaystyle \frac{dN_1}{dt} =
\ok{+c_1N_1}
\ok{-c_2N_1^2}
\ok{-c_3N_1P}
\ok{-c_4N_1Q}$

$\displaystyle \frac{dN_2}{dt} =
\ok{+c_5N_2}
\ok{-c_6N_2^2}
\ok{-c_7N_2P}
\ok{-c_8N_2Q}$

$\displaystyle \frac{dP}{dt} =
\ok{-c_9P}
\ok{+c_{10}N_1P}
\ok{+c_{11}N_2P}
\ok{+c_{12}PQ}$

$\displaystyle \frac{dQ}{dt} =
\ok{-c_{13}Q}
\ok{+c_{14}N_1Q}
\ok{+c_{15}N_2Q}
\ok{-c_{16}PQ}$
& 1.00 & 1.00 & 1.00 & 1.00 & 1.00 & 1.00 & 1.00 & 0.98 \\

With data 
& EX 
& IGP 4-species
& $\displaystyle \frac{dx_1}{dt} =
\ok{+1.00x_1}
\ok{-0.003x_1^2}
\ok{-0.012x_1y_1}
\ok{-0.010x_1y_2}$

$\displaystyle \frac{dx_2}{dt} =
\ok{+0.80x_2}
\ok{-0.003x_2^2}
\ok{-0.010x_2y_1}
\ok{-0.012x_2y_2}$

$\displaystyle \frac{dy_1}{dt} =
\ok{-0.30y_1}
\ok{+0.007x_1y_1}
\ok{+0.005x_2y_1}
\ok{+0.00005y_1y_2}$

$\displaystyle \frac{dy_2}{dt} =
\ok{-0.25y_2}
\ok{+0.005x_1y_2}
\ok{+0.006x_2y_2}
\ok{-0.0001y_1y_2}$
& 1.00 & 0.97 & 1.00 & 1.00 & 1.00 & 1.00 & 1.00 & 0.98 \\

\midrule

No data 
& EX 
& Allee cooperative support
& $\displaystyle \frac{dN}{dt} =
\ok{-c_1N}
\bad{-c_2N^2}
\bad{-c_3N^2C}
\ok{-c_4N^3}
\bad{+c_5CN}
\quad [\text{\scriptsize missing: } \miss{+c_6N^3C}]$

$\displaystyle \frac{dC}{dt} =
\ok{+c_7N}
\ok{-c_8C}
\quad [\text{\scriptsize missing: } \miss{-c_9NC}]$
& 1.00 & 0.90 & 0.86 & 0.75 & 0.80 & 0.74 & 0.70 & 0.83 \\

With data 
& EX 
& Allee cooperative support
& $\displaystyle \frac{dN}{dt} =
\ok{-0.03N}
\ok{-1.10N^3}
\bad{+1.32CN}
\quad [\text{\scriptsize missing: } \miss{+N^2},\ \miss{+N^2C},\ \miss{+N^3C}]$

$\displaystyle \frac{dC}{dt} =
\ok{+0.66N}
\ok{-0.73NC}
\ok{-0.04C}$
& 1.00 & 0.95 & 0.83 & 0.71 & 0.77 & 0.85 & 0.75 & 0.85 \\

\midrule

No data 
& EX 
& FitzHugh--Nagumo adaptation
& $\displaystyle \frac{dv}{dt} =
\ok{+c_1v}
\ok{-c_2v^3}
\ok{-c_3w}
\ok{-c_4a}$

$\displaystyle \frac{dw}{dt} =
\ok{+c_5v}
\ok{-c_6w}$

$\displaystyle \frac{da}{dt} =
\ok{-c_7a}
\bad{+c_8v}
\quad [\text{\scriptsize missing: } \miss{+c_9v^2}]$
& 1.00 & 0.95 & 0.88 & 0.88 & 0.88 & 0.93 & 0.87 & 0.95 \\

With data 
& EX 
& FitzHugh--Nagumo adaptation
& $\displaystyle \frac{dv}{dt} =
\ok{-2.90w}
\bad{+0.30v^2}
\bad{-0.17A^3}
\bad{-0.07w^3}
\quad [\text{\scriptsize missing: } \miss{+v},\ \miss{-\frac{1}{3}v^3},\ \miss{-\chi A},\ \miss{+I}]$

$\displaystyle \frac{dw}{dt} =
\ok{+0.06v}
\ok{-0.08w}
\bad{+0.03v^2}
\quad [\text{\scriptsize missing: } \miss{+c}]$

$\displaystyle \frac{dA}{dt} =
\ok{-0.0005A}
\bad{-0.018v}
\quad [\text{\scriptsize missing: } \miss{+\rho v^2}]$
& 1.00 & 0.92 & 0.44 & 0.50 & 0.47 & 0.69 & 0.64 & 0.45 \\

\midrule

No data 
& DI 
& Misinformation dynamics
& $\displaystyle \frac{dM}{dt} =
\ok{+c_1M}
\ok{-c_2MT}
\ok{-c_3MR}
\bad{-c_4M^2}$

$\displaystyle \frac{dT}{dt} =
\ok{-c_5T}
\ok{-c_6MT}
\bad{-c_7M}
\bad{-c_8R}$

$\displaystyle \frac{dR}{dt} =
\ok{-c_9R}
\ok{-c_{10}MR}
\bad{-c_{11}M}
\bad{+c_{12}T}$
& 1.00 & 1.00 & 0.58 & 1.00 & 0.74 & 0.67 & 1.00 & 0.74 \\

With data 
& DI 
& Misinformation dynamics
& $\displaystyle \frac{dM}{dt} =
\ok{+0.12M}
\ok{-0.12MT}
\ok{-0.05CM}$

$\displaystyle \frac{dT}{dt} =
\ok{+0.01}
\ok{-0.01T}
\ok{-0.01MT}$

$\displaystyle \frac{dC}{dt} =
\ok{+0.02}
\ok{-0.01C}
\ok{-0.02CM}$
& 1.00 & 1.00 & 1.00 & 1.00 & 1.00 & 1.00 & 1.00 & 0.98 \\

\midrule

No data 
& DI 
& Chronic wound
& $\displaystyle \frac{dM}{dt} =
\ok{+c_1DM}
\ok{-c_2IM}
\ok{-c_3M}$

$\displaystyle \frac{dI}{dt} =
\ok{+c_4M}
\ok{-c_5I}
\ok{+c_6D}
\quad [\text{\scriptsize missing: } \miss{-c_7MI},\ \miss{-c_8DI}]$

$\displaystyle \frac{dD}{dt} =
\ok{-c_9D}
\ok{+c_{10}M}
\ok{+c_{11}I}
\quad [\text{\scriptsize missing: } \miss{-c_{12}MD},\ \miss{-c_{13}ID}]$
& 1.00 & 0.96 & 1.00 & 0.69 & 0.82 & 0.69 & 0.69 & 0.90 \\

With data 
& DI 
& Chronic wound
& $\displaystyle \frac{dB}{dt} =
\ok{+0.001B}
\bad{+0.02BI}
\bad{-0.32B^2}
\quad [\text{\scriptsize missing: } \miss{-BH}]$

$\displaystyle \frac{dI}{dt} =
\bad{+1.65B^2}
\quad [\text{\scriptsize missing: } \miss{+1},\ \miss{+B},\ \miss{-BI},\ \miss{-H},\ \miss{-I},\ \miss{+HI}]$

$\displaystyle \frac{dH}{dt} =
\ok{-0.76BH}
\ok{-0.75HI}
\bad{+0.38I}
\bad{-0.89I^2}
\bad{+1.81BI}
\quad [\text{\scriptsize missing: } \miss{+1},\ \miss{-H}]$
& 1.00 & 0.89 & 0.44 & 0.36 & 0.40 & 0.46 & 0.48 & 0.57 \\

\end{longtable}
\end{landscape}

Next, we performed an ablation study on three representative benchmark problems. Recall, the \textit{MEDA} consists of exploratory data analysis and SINDy-based \cite{jiang2021modeling,brunton2016discovering} initialization, literature retrieval, formalization of variables, constraints and candidate seeds, validation, genetic search, numerical data fitting, and final evaluation. From here, we compared the full system against five ablated configurations. In the constraints-only setting, we evaluated a pretrained-formalizer variant without literature retrieval. In the with-data setting, we evaluated four variants: a no-EDA variant in which literature retrieval and formalization were retained but exploratory analysis and SINDy were removed; a no-literature variant in which EDA/SINDy were retained but the formalizer was driven only by data-derived structure; a pretrained-formalizer variant in which both EDA and literature were removed but the pretrained formalizer was retained; and a pure-SINDy variant in which SINDy was used directly without literature, formalization, or explicit constraints.

Table~\ref{tab:ablation_summary} summarizes the performance across the three with-data benchmark problems depended mainly on knowledge-guided formalization rather than numerical fitting alone. The full system recovered the correct term structure in all cases, with average term-level \(F_1=1.00\) and plausibility \(S_{\mathrm{plaus}}=0.967\). Removing EDA/SINDy had little effect, with the no-EDA variant remaining at \(F_1=1.00\) and plausibility (0.969). The pretrained-formalizer variant also stayed near ceiling, despite removing both EDA and literature retrieval, with \(F_1=1.00\) and plausibility (0.983). In contrast, data-driven structure discovery degraded performance. The no-literature variant, which relied on EDA/SINDy-derived formalization, reduced average plausibility to (0.700) and term-level \(F_1\) to (0.732). The strongest degradation occurred in misinformation dynamics, where plausibility dropped from (0.98) to (0.59) and term-level \(F_1\) from (1.00) to (0.50), despite good trajectory fitting. Thus, numerical fit alone did not guarantee recovery of the correct symbolic mechanism. The pure-SINDy variant showed the largest degradation. Without literature, formalization, or explicit constraints, average plausibility dropped to (0.603), term-level \(F_1\) to (0.541), and term precision to (0.382). SINDy often recovered dominant driver terms, but introduced many false-positive terms, especially in conserved or coupled equations. In SIR-type systems, recovery equations became unstable without conservation constraints; in misinformation dynamics, spurious terms remained in the trust and moderation-capacity equations. The constraints-only pretrained ablation further showed that data anchoring is useful when a familiar model must be adapted to a more specific variant. Without data, pretrained formalization stayed close to the full constraints-only system for misinformation dynamics, with plausibility (0.73) versus (0.74). However, in age-structured SIR, plausibility dropped from (0.964) to (0.744), and term recall fell to (0.571), because the model collapsed the heterogeneous system toward a simpler homogeneous SIR structure.
\begin{table}[!ht]
\centering
\caption{Ablation study for the with-data setting, divided by experimental setting. IR: information retrieval; EX: reasoning-based extrapolation; DI: discovery.}
\label{tab:ablation_summary}
\resizebox{\textwidth}{!}{
\begin{tabular}{llccccccc}
\hline \hline
\textbf{Setting} & \textbf{System variant} & \textbf{Plaus.} & \textbf{Term \(F_1\)} & \textbf{Term P} & \textbf{Term R} & \textbf{Constr. \(F_1\)} & \textbf{Mean incl.} & \textbf{Mean TED} \\
\hline\hline

\textbf{IR} & Full system & 0.980 & 1.000 & -- & -- & 1.000 & 1.000 & -- \\
& No EDA/SINDy & 0.986 & 1.000 & 1.000 & 1.000 & 1.000 & 1.000 & 0.000 \\
& No literature & 0.780 & 0.800 & 0.667 & 1.000 & 0.957 & 1.000 & 0.254 \\
& Pretrained formalizer & 1.000 & 1.000 & 1.000 & 1.000 & 1.000 & 1.000 & 0.000 \\
& Pure SINDy & 0.760 & 0.667 & 0.500 & 1.000 & n/a & 1.000 & 0.250 \\

\hline

\textbf{EX} & Full system & 0.940 & 1.000 & -- & -- & 1.000 & 1.000 & -- \\
& No EDA/SINDy & 0.982 & 1.000 & 1.000 & 1.000 & 0.960 & 1.000 & 0.000 \\
& No literature & 0.730 & 0.897 & 0.867 & 0.929 & 0.982 & 0.963 & 0.097 \\
& Pretrained formalizer & 0.980 & 1.000 & 1.000 & 1.000 & 1.000 & 1.000 & 0.000 \\
& Pure SINDy & 0.570 & 0.393 & 0.255 & 0.857 & n/a & 0.926 & 0.438 \\

\hline

\textbf{DI} & Full system & 0.980 & 1.000 & -- & -- & 1.000 & 1.000 & -- \\
& No EDA/SINDy & 0.938 & 1.000 & 1.000 & 1.000 & 1.000 & 1.000 & 0.000 \\
& No literature & 0.590 & 0.500 & 0.368 & 0.778 & 0.667 & 0.833 & 0.450 \\
& Pretrained formalizer & 0.970 & 1.000 & 1.000 & 1.000 & 0.963 & 1.000 & 0.000 \\
& Pure SINDy & 0.480 & 0.563 & 0.391 & 1.000 & n/a & 1.000 & 0.444 \\

\hline\hline
\end{tabular}
}
\end{table}

Finally, we evaluated the robustness of \textsc{MEDA} to corrupted or incomplete data. This analysis focused on three representative benchmark problems, one from each difficulty tier: classical SIR for IR, age-structured SIR for EX, and misinformation dynamics for DI. We considered three data-level perturbations: proportional Gaussian measurement noise, sparse sampling of the observed trajectory, and hidden-variable settings in which one state variable was removed from the observed data. The prompt-perturbation axis was not included in this analysis and is treated separately from the present robustness experiments. For the measurement-noise experiments, we used a fast-path evaluation in which the knowledge-derived problem specification was held fixed, and only the genetic search and data-fitting stages were re-run on corrupted trajectories. This isolates the robustness of the symbolic search and numerical fitting stages from stochastic variation in the LLM-based literature and formalization stages. For the hidden-variable experiments, we re-ran the full pipeline because removing a state variable changes the admissible ODE system and requires the constraints and seed equations to be re-derived.

Table~\ref{tab:robustness_noise} summarizes the robustness to proportional Gaussian measurement noise in terms of \(F_1\) under increasing noise levels. The classical SIR model retained perfect term-level recovery through 30\% proportional Gaussian noise and degraded only at 40--60\%, where term \(F_1\) decreased to 0.73. Importantly, this degradation reflected reduced precision rather than loss of the true mechanism: the model continued to recover all reference terms but added spurious cross-terms under high noise. The age-structured SIR system was similarly robust, with term \(F_1 \geq 0.90\) across all tested noise levels. At 30\% noise, the correct term structure was recovered, although the best-fitted coefficients produced a divergent trajectory during integration; this case is therefore marked as structurally correct but numerically unstable. In contrast, misinformation dynamics showed little additional degradation with noise because its recovery was already limited in the clean condition. Its term \(F_1\) remained within a narrow range of 0.57--0.67 across all noise levels, suggesting that the bottleneck was structural identifiability rather than measurement noise.

\begin{table}[!ht]
\centering
\caption{Robustness to proportional Gaussian measurement noise. Values report term-level \(F_1\) under increasing noise levels.}
\label{tab:robustness_noise}
\resizebox{\textwidth}{!}{
\begin{tabular}{lccccccc}
\hline\hline
\textbf{Problem} & \textbf{0\%} & \textbf{5\%} & \textbf{10\%} & \textbf{20\%} & \textbf{30\%} & \textbf{40\%} & \textbf{60\%} \\
\hline\hline
SIR epidemic (IR) & 1.00 & 1.00 & 1.00 & 1.00 & 1.00 & 0.73 & 0.73 \\
Age-structured SIR (EX) & 1.00 & 1.00 & 1.00 & 1.00 & 1.00 & 0.90 & 1.00 \\
Misinformation dynamics (DI) & 0.67 & 0.63 & 0.63 & 0.63 & 0.57 & 0.67 & 0.63 \\
\hline\hline
\end{tabular}
}
\vspace{0.5em}
\begin{flushleft}
\end{flushleft}
\end{table}

Table~\ref{tab:robustness_hidden}, removing the recovered compartment \(R\) from classical SIR had no effect on observed-state recovery, because \(R\) does not feed back into the dynamics of \(S\) or \(I\). Similarly, dropping \(R_O\) from age-structured SIR preserved perfect term-level recovery and high plausibility, since this variable enters only weakly through the force-of-infection normalization. In contrast, dropping the moderation-capacity variable \(C\) from misinformation dynamics reduced variable \(F_1\) to 0.80 and term \(F_1\) to 0.83. This is expected because \(C\) directly suppresses misinformation through the \(-CM\) interaction. When \(C\) is latent, the system can recover the observed misinformation-growth and trust-coupling terms, but cannot explicitly represent the missing moderation-suppression mechanism. I.e., sparse sampling had little effect in this synthetic benchmark since removing 25--75\% of the time points from the noise-free trajectories left the recovered terms unchanged for both SIR systems, and misinformation dynamics fluctuated only within its baseline recovery range. 

\begin{table}[!ht]
\centering
\caption{Robustness to hidden variables. Values report performance after one state variable is removed from the observed data and the model is re-derived.}
\label{tab:robustness_hidden}
\resizebox{\textwidth}{!}{
\begin{tabular}{lccccc}
\hline\hline
\textbf{Problem} & \textbf{Dropped state} & \textbf{Variable \(F_1\)} & \textbf{Term \(F_1\)} & \textbf{Inclusion} & \textbf{Plaus.} \\
\hline\hline
SIR epidemic (IR) & \(R\) & 1.00 & 1.00 & 1.00 & 1.00 \\
Age-structured SIR (EX) & \(R_O\) & 1.00 & 1.00 & 1.00 & 0.98 \\
Misinformation dynamics (DI) & \(C\) & 0.80 & 0.83 & 0.83 & 0.89 \\
\hline\hline
\end{tabular}
}
\end{table}

\section{Discussion}
In this study, we introduce \textsc{MEDA}, an agentic framework for biological ODE discovery that couples LLM-based scientific reasoning with constrained SR. The central finding is that LLM agents are most useful in this setting not as direct equation generators, but as mechanisms for translating incomplete biological knowledge into variables, constraints, candidate model structures, and critique criteria that can guide symbolic search. Across canonical retrieval tasks, reasoning-based extrapolation tasks, and discovery-oriented tasks, \textsc{MEDA} often recovered the correct state space and a plausible mechanistic skeleton even when the final expression was not exactly identical to the reference model (Table~\ref{tab:result1}). This supports a view of automated scientific discovery in which the goal is not simply to produce a final symbolic expression, but to construct an auditable chain of mechanistic reasoning around that expression.

The results also show that equation discovery in biological systems is limited less by coefficient fitting than by structural ambiguity. In several cases, fitting data improved the recovered model. Nevertheless, the benefit of data was not monotonic. In some cases, the data-anchored mode produced trajectory-compatible but mechanistically incorrect equations, missing defining couplings while introducing spurious terms. This behavior is consistent with a broader challenge in biological modeling where finite, noisy, and partially observed trajectories can support multiple plausible dynamical explanations, and parameter or structural non-identifiability can persist even when trajectory fit appears strong \cite{gutenkunst2007sloppy,raue2009structural}. Thus, the relevant test for a discovered biological model is not only whether it fits observed data, but whether it recovers the mechanisms required to explain the system.

The ablation analysis clarifies which parts of the pipeline are responsible for this behavior. As shown in Table~\ref{tab:ablation_summary}, the full system and the variants that retained a knowledge-guided formalizer remained near the performance ceiling, whereas data-only variants degraded sharply. Removing EDA/SINDy had little effect, and the pretrained-formalizer variant also remained strong, indicating that for named and well-studied systems, literature retrieval and pretrained biological knowledge can act as alternative sources of mechanistic prior. By contrast, the no-literature data-driven formalizer and the pure-SINDy variant produced substantially lower plausibility and term precision. This result reinforces a key point for machine learning approaches to scientific discovery: sparse regression and numerical optimization can identify candidate terms, but without an external mechanistic prior they may preserve false-positive structures that fit the data but do not express the correct biology \cite{brunton2016sindy,champion2019data}. In this sense, the formalizer and constraint-generation stages are load-bearing components rather than explanatory decoration.

The robustness analysis further qualifies this conclusion. As shown in Table~\ref{tab:robustness_noise}, the system was not uniformly sensitive to data corruption. For the IR and EX related tasks, term-level recovery remained high under substantial proportional measurement noise. When degradation occurred, it was mainly due to reduced precision rather than loss of the true mechanism, meaning that the correct terms were still present but accompanied by spurious additions. By contrast, the DI task did not show a strong noise-response curve because its clean-data recovery was already imperfect, suggesting that the limiting factor was structural identifiability rather than measurement noise. The hidden-variable experiments, as indicated by Table~\ref{tab:robustness_hidden}, show a similar pattern. Dropping terminal or weakly coupled variables, did not harm recovery, whereas removing variables in the DI task reduced both variable and term recovery. Thus, the main robustness limitation is not generic sensitivity to noisy or sparse data, but the loss of mechanistically important latent drivers and the resulting ambiguity in the admissible dynamics. From these results, one can conclude that knowledge-guided equation discovery should be evaluated not only by fit quality, but also by whether the available variables are sufficient to identify the mechanisms being modeled \cite{gutenkunst2007sloppy,raue2009structural}.

These findings help position \textsc{MEDA} relative to recent work on SR and ASD, and LLM-assisted science. Classical SR and SINDy-like methods search for compact equations directly from data \cite{schmidt2009distilling,brunton2016sindy,makke2024srreview}. Hybrid scientific machine-learning frameworks combine mechanistic differential equations with flexible learned components, often improving predictive accuracy while preserving some physical structure \cite{rackauckas2020universal,giampiccolo2024robust}. LLM-based equation discovery extends this direction by injecting natural-language priors or programmatic search heuristics into SR \cite{shojaee2025llmsr,taskin2026knowledge}. \textsc{MEDA} differs by treating the intermediate scientific artifacts themselves as part of the discovery object. This is particularly important for biology, where prior knowledge is often qualitative, incomplete, and distributed across the literature rather than expressible as exact conservation laws or symmetries.

Several limitations remain. First, the benchmark is small relative to the diversity of biological ODE modeling. Although it spans population dynamics, epidemiology, neural excitability, ecological interactions, misinformation dynamics, and wound healing, it does not cover the full range of biochemical networks, spatial dynamics, stochastic processes, delay differential equations, or hybrid discrete-continuous models used in biology. Second, the current search grammar emphasizes polynomial ODEs. This makes the evaluation controlled, but it excludes important biological structures such as Hill kinetics, saturating functional responses, delays, thresholds, and stochastic effects. Finally, the plausibility score is expert-based and therefore necessarily subjective, even though it was averaged across five experienced biomathematicians. 

Taken jointly, \textsc{MEDA} suggests a path toward AI systems that assist scientific modeling by making mechanistic assumptions explicit, testable, and revisable. Such systems are unlikely to replace expert modelers, but they may help accelerate the iterative process of proposing, criticizing, and refining mathematical hypotheses.

\section{Methods}
In this section, we outline the experimental design and evaluation metrices. The full technical description of how \textit{MEDA} is structured is provided in the appendix.

\subsection{Experimental design}
We evaluated \textsc{MEDA} across three increasingly difficult settings: information retrieval, reasoning-based extrapolation, and open-ended discovery. The full descriptions of all benchmark systems, including the reference equations, variable definitions, and parameter interpretations, are provided in the Appendix.

Simply put, in the information-retrieval setting, the goal was to test whether the system could recover standard ODE models that are widely documented in the literature. We used five canonical systems: logistic population growth, Lotka--Volterra predator--prey dynamics, the classical SIR epidemic model, a fixed-threshold Allee-effect population model, and the FitzHugh--Nagumo model of neural excitability. These tasks primarily evaluate whether the agent can identify known state variables, constraints, and equation structures from background knowledge. In the reasoning-based extrapolation setting, the goal was to test whether the system could adapt familiar model families to previously unseen mechanistic variants. We therefore constructed five extensions of canonical biological models: a four-species intraguild predation system, logistic growth under two antagonistic biological pressures, an age-structured SIR model with waning immunity only in the young group, an Allee-effect model with dynamic cooperative support, and a FitzHugh--Nagumo-type model with slow adaptation. These systems are grounded in established biological mechanisms, but the exact equations were designed for this study and were not intended to be direct retrieval targets. In the discovery setting, the goal was to test whether the system could construct plausible mechanistic ODEs for less standardized phenomena. We considered two systems: misinformation dynamics with trust erosion and moderation fatigue, and chronic wound progression driven by microbial burden, inflammation, and tissue-barrier integrity. In these tasks, the reference equations serve as expert-defined mechanistic targets rather than canonical models from the literature. All tasks were evaluated in a constraints-only mode, in which no time-series data were provided, and, when applicable, in a data-anchored mode, in which simulated trajectories were supplied for fitting. Ground-truth equations were withheld from all discovery agents and were used only after model generation for quantitative evaluation.

\subsection{Evaluation metrics}
We evaluated \textsc{MEDA} at multiple levels of the discovery process: state-variable recovery, constraint recovery, term-level equation recovery, whole-expression similarity, and expert-assessed biological plausibility. All metrics were computed separately for the information-retrieval, reasoning-based extrapolation, and discovery settings. Full metric definitions and equations are provided in Appendix.

Variable recovery measures whether the system identified the correct dynamical state space. Predicted variables were matched to reference variables after semantic normalization, so that notationally different but biologically equivalent variables were treated as correct. We report variable precision, recall, and \(F_1\). Constraint performance measures whether the intermediate biological and mathematical restrictions generated by the system were correct, relevant, and specific to the task. These constraints include, for example, non-negativity, conservation of total population in closed compartmental systems, boundedness under resource limitation, and correct signs of interactions such as predation, infection, recovery, or inhibition. Generated constraints were compared with an expert-consensus reference set, and we report constraint precision, recall, \(F_1\), and accuracy. Term-level performance evaluates whether the generated ODEs recovered the correct mathematical building blocks. Each right-hand side was decomposed into additive symbolic terms after normalization of parameter names, scalar coefficients, and algebraically equivalent forms. Terms were matched only when they appeared in the correct state equation with the correct structural role. We report term precision, recall, \(F_1\), and term-level accuracy over a task-specific admissible term universe. To evaluate whole-equation structure, we also compared predicted and reference expressions as normalized expression trees. We report normalized tree edit distance (TED), where lower values indicate greater symbolic similarity, and an inclusion score, which measures how much of the reference expression appears as a correct substructure within the generated expression. The inclusion score is useful when the model recovers the main mechanism but adds extra terms or uses a slightly different symbolic organization. Finally, biological plausibility was assessed by five independent evaluators, all of whom held a PhD in mathematics and had at least ten years of professional experience. Each evaluator assigned a score between 0 and 1, considering interaction signs, satisfaction of biological and mathematical constraints, mechanistic completeness, interpretability of variables and parameters, and consistency with the qualitative behavior described in the prompt. The final plausibility score was the average of the five evaluator scores.

\section*{Declarations}
\subsection*{Funding}
No funding was received to assist with the preparation of this manuscript.

\subsection*{Conflicts of interest/Competing interests}
The authors have no competing interests to declare that are relevant to the content of this article.

\subsection*{AI usage}
The authors used AI models for a wide range of tasks in this study, including: ideation, code development, experiment design, experiment execution, and initial manuscript draft preparation. All AI outputs were manually checked and validated, and the authors take full responsibility for the final content. 

\subsection*{Data availability}
All data used for this study are provided in the manuscript and the appendix.

\subsection*{Code availability}
The code is freely available in the study's Github repository: \url{https://github.com/kronga/MEDA} 

\subsection*{Author Contribution}
David Krongauz: Methodology, Software, Formal analysis, Investigation, Data Curation, Writing - Review \& Editing, Visualization. \\ Arad Zulti: Methodology, Software, Formal analysis, Writing - Review \& Editing, Visualization. \\ Eran Segal: Validation, Resources, Writing - Review \& Editing, Supervision. \\ Teddy Lazebnik: Conceptualization, Methodology, Validation, Formal analysis, Data Curation, Writing - Original Draft, Visualization, Project administration.

\bibliography{biblio.bib}

\end{document}